\documentclass[12pt]{article}
\usepackage[english]{babel}
\usepackage[utf8x]{inputenc}
\usepackage{amsmath}
\usepackage{graphicx}
\usepackage{etoolbox}
\usepackage{changepage}
\usepackage{titlesec}
\usepackage[parfill]{parskip}
\usepackage[margin=1in]{geometry}
\usepackage{times}
\usepackage{float}
\usepackage[numbers,super]{natbib}
\usepackage{lipsum} % Package to generate dummy text throughout this template. Remove for real use.
\usepackage{mathptmx}

\titleformat{\subsection}
  {\normalfont\fontsize{12}{15}\bfseries}{}{0em}{}

\usepackage{adjustbox}
\usepackage{url}
\titleformat*{\section}{\normalsize\bfseries} % Makes section titles 12 pt font

\title{\large \textbf{Undergraduate Robotics Education with General Instructors using a Student-Centered Personalized Learning Framework}} % using \large makes the title approximately 14 pt.
\author{}
\date{} % This leaves the date blank.

\makeatletter % This gets the margins for the title set.
\patchcmd{\@maketitle}{\begin{center}}{\begin{adjustwidth}{0.5in}{0.5in}\begin{center}}{}{}
\patchcmd{\@maketitle}{\end{center}}{\end{center}\end{adjustwidth}}{}{}
\makeatother

\usepackage[dvipsnames]{xcolor}
\definecolor{mypink}{RGB}{0, 0, 0}
\newcommand{\ruic}[1]{\textcolor{mypink}{{#1}}}

\usepackage{authblk}
\author[1]{Rui Wu}
\author[2]{David J Feil-Seifer}
\author[2]{Ponkoj C Shill}
\author[2]{Hossein Jamali}
\author[2]{Sergiu Dascalu}
\author[2]{Fred Harris}
\author[3]{Laura Rosof}
\author[3]{Bryan Hutchins}
\author[4]{Marjorie Campo Ringler}
\author[5]{Zhen Zhu}
\affil[1]{Department of Computer Science, East Carolina University}
\affil[2]{Department of of Computer Science \& Engineering, University of Nevada, Reno}
\affil[3]{SERVE Center | Early College Research Center }
\affil[4]{Department of Educational Leadership, East Carolina University}
\affil[5]{Department of Engineering, East Carolina University}

\begin{document}
\raggedright
\maketitle
\thispagestyle{empty}
\pagestyle{empty}

%----------------------------------------------------------------------------------------
%  PAPER CONTENTS
%----------------------------------------------------------------------------------------
\section*{Abstract}
% possible division
% NSF Grantees Poster Session
% Engineering Technology Division (ETD) 
% New Engineering Educators (NEE) Division
% For authors: https://www.asee.org/events/Conferences-and-Meetings/2024-Annual-Conference/Paper-Management/2023-Authors
% Length: There is no set limit for the number of pages a paper can or must be.
% sample paper: https://peer.asee.org/cultural-dimensions-in-academic-disciplines-a-comparison-between-ecuador-and-the-united-states-of-america
Recent advancements in robotics, including applications like self-driving cars, unmanned systems, and medical robots, have had a significant impact on the job market. On one hand, big robotics companies offer training programs based on the job requirements. However, these training programs may not be as beneficial as general robotics programs offered by universities or community colleges. On the other hand, community colleges and universities face challenges with required resources, especially qualified instructors, to offer students advanced robotics education. Furthermore, the diverse backgrounds of undergraduate students present additional challenges. Some students bring extensive industry experiences, while others are newcomers to the field. To address these challenges, we propose a student-centered personalized learning framework for robotics. This framework allows a general instructor to teach undergraduate-level robotics courses by breaking down course topics into smaller components with well-defined topic dependencies, structured as a graph. This modular approach enables students to choose their learning path, catering to their unique preferences and pace. Moreover, our framework's flexibility allows for easy customization of teaching materials to meet the specific needs of host institutions. In addition to teaching materials, a frequently-asked-questions document would be prepared for a general instructor. If students' robotics questions cannot be answered by the instructor, the answers to these questions may be included in this document. For questions not covered in this document, we can gather and address them through collaboration with the robotics community and course content creators. Our user study results demonstrate the promise of this method in delivering undergraduate-level robotics education tailored to individual learning outcomes and preferences.

%------------------------------------------------

\section{Introduction}
\ruic{University and community college education, aimed at workforce preparation, can sometimes face limitations in course and degree availability. While courses related to robotics and advanced manufacturing are often listed in catalogs, students may encounter difficulty enrolling due to a shortage of robotics instructors, as observed in our collaborative work with North Carolina and Nevada community colleges and universities. This shortage, based on our practical experience and literature survey, stems primarily from two issues: instructor expertise and the diverse backgrounds of undergraduate students}:

\ruic{- Issue 1: The number of robotics faculty available for teaching are not enough to meet the instructional need. For community colleges in particular, this can be challenging for multiple reasons~\cite{Quinterno_2020}.
% , such as less salary, lack of academic ranks (i.e., assistant professor, associate professor, and professor), and lack of tenure protections. 
This issue can be more serious in rural areas~\cite{murray2007recruiting,cejda2011hiring,dias2007undergraduate,rosenblatt2000designing}. 
% Based on the news~\cite{Krupnick_2018} and research results~\cite{murray2007recruiting,cejda2011hiring}, it is challenging for rural community colleges to recruit and retain qualified instructors. Some rural community colleges lower the hiring requirements to recruit industry people but usually they do not have any teaching experience. Universities can have the same issue~\cite{dias2007undergraduate,rosenblatt2000designing}. 
Emerging robotics topics, such as self-driving cars and collaborative robots, can be interdisciplinary including physics, mathematics, and computer science. Furthermore, robotics education priorities can change rapidly, even every year. It can be very challenging for a regular instructor without robotics expertise to teach these robotics topics. One way to address this issue could be to reduce the entry barrier for robotics instruction. But this can have a negative impact on the quality of robotics courses. Hiring a capable robotics instructor is challenging. Training a robotics instructor is difficult too. Instructors usually have heavy teaching loads and have limited time to learn new skills and prepare new courses. 
% For example, an instructor is required to teach three or four courses per semester at the ECU Computer Science department. 
An instructor may spend multiple years to be ready to teach emerging robotics topics.}

\ruic{- Issue 2:} Undergraduate students can be very different. Some students may have abundant industry experience and come back to school to learn a specific skill. Others may be new students that have never taken any undergraduate courses. In addition, robotics requires a large combination of prerequisite skills, from computer programming and mathematics, to physics and engineering. Students may engage differently with the material required for mastery in robotics. Therefore, their learning pace and preferences can be very different. “How to efficiently educate students with different backgrounds” is also a challenging issue. Compared to the traditional “instructor-centered” pedagogy, a “student-centered” pedagogy allows a student to have more freedom and can create a stronger learning environment.

To solve the undergraduate robotics education issues mentioned above, we propose a student-centered personalized learning framework that allows an instructor without robotics expertise to teach undergraduate students robotics knowledge. “Student-centered” means that an individual student will decide his/her learning path and pace, which is different from the traditional “instructor-centered” teaching in which an instructor controls the teaching flow and speed~\cite{rasmussen1956evaluation,hovey2019frequency}. The “robotics knowledge” should fill the gap between the current curriculum commonly taught in the academic world and the requirements from local robotics companies. 

Interactive System for Personalized Learning (ISPeL)~\cite{ispel_website} has been implemented based on our proposed learning framework. Feedback from over 100 students on ISPeL has been collected, and the results of the user study show that our proposed framework is promising for enhancing undergraduate education. Students have found it more convenient to understand how topics are connected and to review the knowledge they have acquired through ISPeL.

In the rest of this paper, Section \textbf{Related Work} discusses related personalized learning pedagogy and e-learning platforms; Section \textbf{Methods} illustrates our proposed framework and a personalized learning platform based on the proposed framework; Section \textbf{User Study Results and Discussion} presents our user study results and discussions.
%------------------------------------------------

\section{Related Work}
\label{s:related_work}
\subsection{Personalized Learning Pedagogy}
In adult learning theory, Knowles~\cite{knowles1984adult} presents the andragogy theory, which holds significant relevance to personalized learning pedagogy, particularly in the online learning domain. Malcolm Knowles' adult learning theory remains a cornerstone in higher education, shaping curriculum and pedagogical approaches. Aligned with adult learning theory, personalized learning pedagogies address the assumption that learners, especially college students, have an inherent need to comprehend the purpose behind their learning endeavors, whether revisiting familiar concepts or exploring new domains.

Within the general framework of personalized learning, activities grounded in real-life scenarios enhance student engagement, particularly evident in fields such as robotics and computer science, where problem-solving and scenario-based learning align with adult learning principles. Derived from andragogy, four key principles characterize adult-centered instruction and learning~\cite{cross1981adults,knowles1984adult,ota2006training}: relevance to assignments, encouragement of critical and reflective thinking, acknowledgment and utilization of personal knowledge and experiences, and self-paced learning.

Effective personalized learning pedagogies incrementally build success, validate achievement, provide constructive feedback for learning from errors~\cite{wanichsan2021enhancing}, and progressively increase rigor. Considering the diversity of learners’ experiences, personalized learning pedagogies recognize and leverage varied levels of prior knowledge in an era marked by easy access to information and technology. In the contemporary educational landscape, where information is readily available, professors transition from being gatekeepers of knowledge to facilitators of learning. Personalized learning empowers students to create their learning paths, fostering ownership of their educational journey~\cite{shaver2017added}.

\ruic{In the context of college teaching, personalized learning pedagogies offer a viable option, especially for topics that are needed but an institution may not employ faculty with the specific expertise. A personalized learning pedagogy has the potential to tailor learning pathways to individual learning styles and needs, granting learners autonomy and the ability to direct their educational pursuits. An online personalized learning pedagogy, mediated by technology, can potentially bridge gaps in accessibility and provide diverse learning paths and practical experiences to support individual concept mastery~\cite{knowles1984adult,xiao2018personalized}.}

\subsection{E-learning and Personalized Learning Platforms}
\ruic{Traditional computer science undergraduate students utilize technology for learning and receiving feedback~\cite{king2017reimagining}. They constantly receive new information and feedback digitally. It is important that instructors use technology to enhance student learning and engagement as they transition from instructor-centered teaching to learner-centered education. This paper presents a personalized learning pathway that embraces online learning and capitalizes on learners' preferences.}

E-learning at universities is common. Instructor-centered e-learning is often found in traditional learning management systems (LMS) such as Moodle, Canvas, or Blackboard. The content in instructor-centered LMS is often static. In contrast, e-learning systems are considered learner-centered. These learner-centered systems are characterized by personalized learning and are adaptable to the student's needs. E-learning systems are becoming increasingly more available and preferred by college students~\cite{xie2019trends}.  

\ruic{Personalized learning is defined as instruction designed to optimize the learning needs of students, with systems in place for students to learn at their own pace~\cite{king2017reimagining}. Systems developed to personalize learning incorporate learning activities that appeal to the learner's interests. One such system used for teaching computer science program development is APELS, an adaptable and personalized e-learning system~\cite{aeiad2019adaptable}. APELS is innovative as it employs natural language processing to validate content against learning outcomes and leverages the Web to source resources for the learner. Content in systems like APELS is both general and learner-centered.}

\ruic{Sequencing curriculum in an e-learning personalized platform is challenging when considering the level of difficulty of a concept and its relationship with other concepts. The sequencing of concepts in e-learning varies based on learner preferences, existing knowledge, and their learning goals~\cite{agbonifo2018genetic}. Balancing learner preferences with curriculum sequencing is challenging yet necessary. To achieve this balance, e-learning platforms offering personalized learning should include assessments to determine learners' prior knowledge and abilities. By doing so, the platform can offer the learner a pathway and level of difficulty that aligns with their zone of proximal development (ZPD). ZPD occurs when a learner is assisted in learning new concepts or skills~\cite{vygotsky1978mind}. This assistance is known as scaffolding, providing learners with guidance and activities that facilitate learning. In an e-learning personalized platform, the platform provides the tools and activities to promote ``new'' learning, which occurs within the ZPD. Therefore, the e-learning platform must carefully consider the curriculum sequence to gradually increase difficulty while maintaining engagement. If the level of difficulty increases too quickly, learners may become frustrated. Conversely, if the difficulty level is not challenging enough or the process does not appeal to their interests, learners may become disengaged. ZPD is a dynamic concept that can be addressed in e-learning platforms. Key factors that make ZPD a learning target include: 1) a variety of concepts with increasing difficulty levels in the e-learning platform curriculum; 2) activities providing scaffolds that guide learners through learning levels; 3) interactions on the e-platform allowing learners to work on their own knowledge and skills. Our proposed methods adhere to these three key factors.}

This paper introduces the Interactive System for Personalized Learning (ISPeL) system, designed to cater to personalized learning based on our proposed student-centered personalized learning framework within e-learning systems, specifically in selected topics of computer science. ISPeL precisely outlines its topics and tailors them to individual student interests. A crucial distinction lies in the fact that, while the curriculum in this e-learning system is determined by a college course, the learning of the content is personalized (more details are introduced in Section~\ref{s:ISPeL} ISPeL). This system acknowledges the uniqueness of each learner and recognizes that a 'one size fits all' approach is neither appealing nor effective in computer science education.

\section{Methods}
\label{s:methods}
\subsection{Student-Centered Personalized Learning Framework}

\begin{figure}
    \centering
    \includegraphics[width=0.65\linewidth]{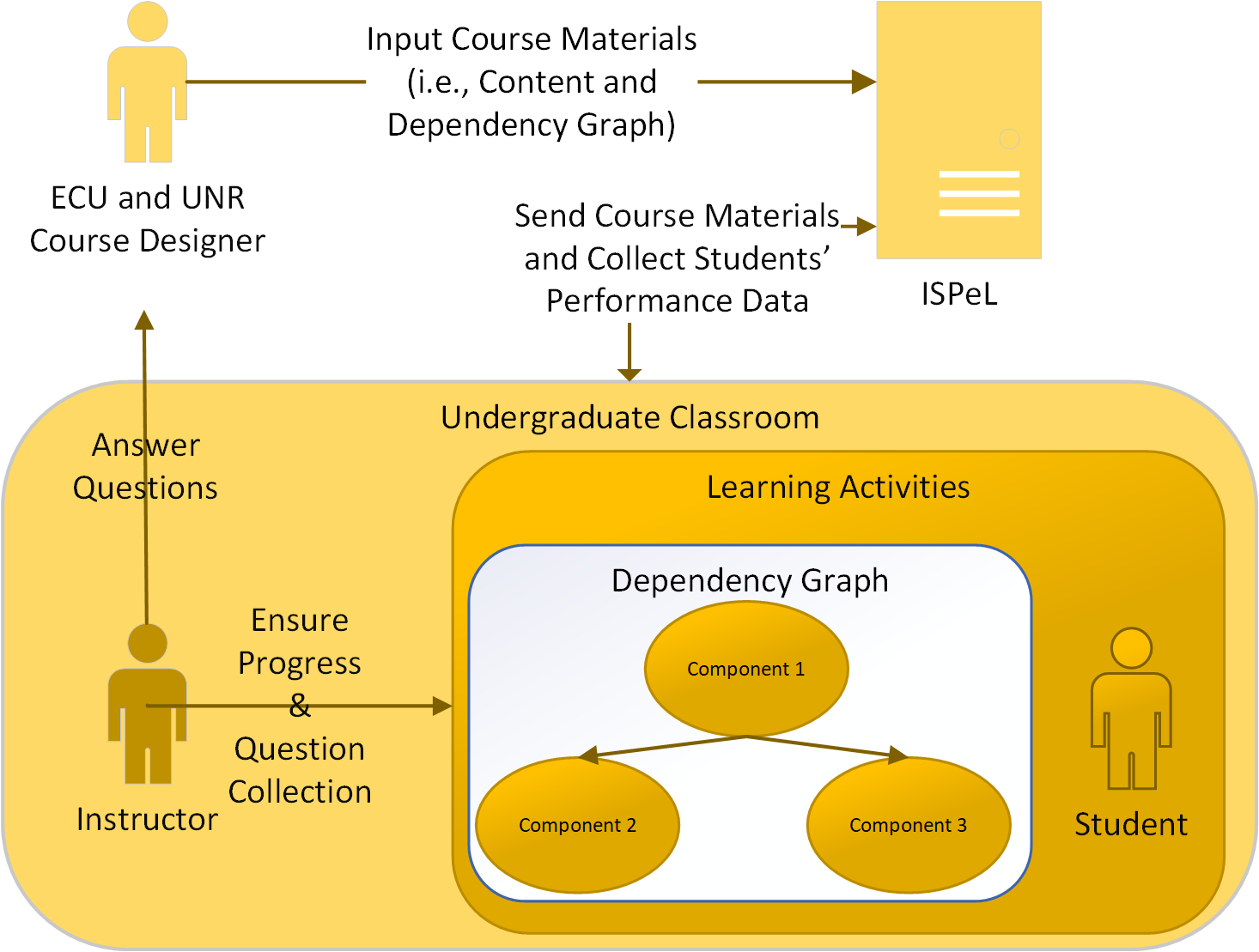}
    \caption{\ruic{Proposed student-centered personalized learning framework: Students are placed at the center, with instructors required to assist rather than lead them. Course materials are developed by faculties from ECU and UNR, encompassing reading materials, hands-on exercises, and dependency graphs (see Figure 2 for more details). Instructors must ensure that students make progress in every class and gather questions from them. Students can work on different topics within the same classroom with different required hardware. Computationally intensive tasks will be carried out on a personalized learning platform server named ISPeL. Additionally, ISPeL will gather students' performance data and track their learning progress.}}
    \label{fig:overall_workflow}
\end{figure}

Self-Determination Theory posits that increasing students’ autonomy can enhance their motivation and engagement~\cite{deci2012self}. However, knowledge should be scaffolded in an intentional manner, as some knowledge dependencies exist across content. In a traditional instructor-centered classroom, what to teach is defined before class and an instructor controls the timing and order of course material. This strategy has some challenges, as a student brings their own unique set of knowledge, interests, and experiences to a classroom. For example, CSCI1010 is an entry-level course for computer science students at East Carolina University. According to our experience, some students enrolled in the course never took any computer science courses. Other students may have learned basic programming knowledge and are familiar with Matlab and R. How to provide a personalized experience (i.e., both learning path and appropriate pace for students with different backgrounds) is a challenge. To tackle this challenge, we propose a student-centered personalized learning framework as Figure~\ref{fig:overall_workflow} shows.

Central to our proposed framework is topic-based authoring, which involves first identifying domains and topics. Topics are small, self-contained, reusable, and context-free content units. Learners may study a topic subject to its prerequisite dependencies. Pre- and post-tests are associated with each domain. Topics can be aggregated to form higher-level learning units, i.e., domains. The ISPeL system keeps track of individual learners’ progress and performance on pre- and post-tests associated with the domains. Learner progress and performance dashboards make it easy to assess learning.

\begin{figure}[h]
\centerline{\includegraphics[width=\columnwidth]{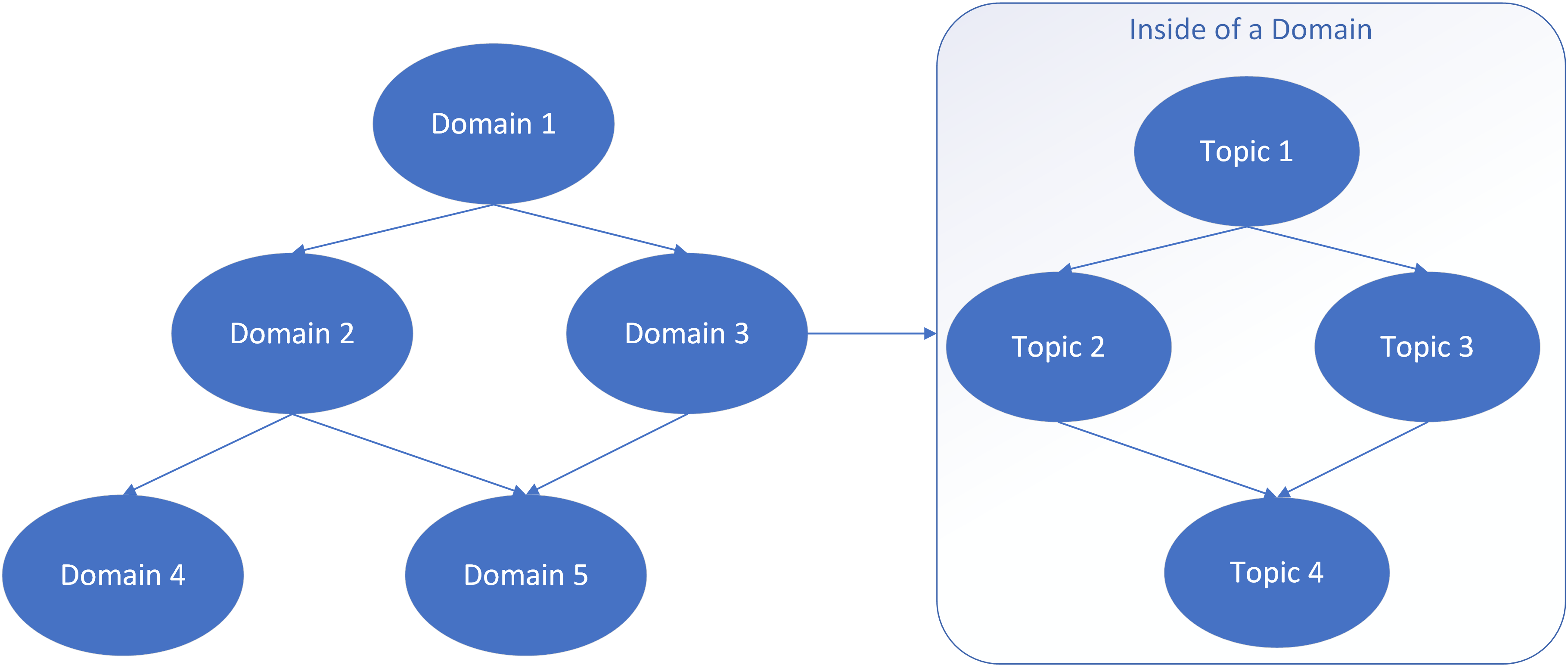}}
\caption{Course domain and domain dependency graph. Domain 1 will be studied first; Domain 5 cannot be reached until a student learns Domains 1, 2, and 3. A student can choose a unique path to learn the required course content. Each domain can include multiple topics, e.g., Domain 3 contains four topics.}
\label{fig:dependency_graph}
\end{figure}

A domain can include multiple topics, visualized using a dependency graph (Fig.~\ref{fig:dependency_graph}). Each topic should stand alone and be not overlap with other topics. Topics should be organized as a Directed Acyclic Graph (Thulasiraman \& Swamy, 1992). For example, how to evaluate a path planning machine learning model performance is an important topic in the self-driving domain. Instead of partially introducing possible evaluation methods in multiple topics, it is more organized to illustrate all popular methods in a standalone course topic. Besides the topics, we also need to define topic dependencies. For each student, the learning order of topics within a domain can be different. Some topics should be learned before other topics. For example, a student should learn how to tune hyperparameters of an image analysis algorithm first, and then he/she can understand how to configure a classification model introduced in the object detection course topic. The dependencies between each course topic should be expressed in the dependency graph structure. 

The vertices represent course content and the edges represent dependencies. The dependencies for both domain and topics should also be defined, but can also be flexible to allow for multiple pathways through the domain. For each student, the order course content is explored can be different. For example, a student should learn basic chemistry principles before understanding mineral composition and physical properties. The dependencies between each course domain and its topics will be expressed in the dependency graph structure (Fig.~\ref{fig:dependency_graph}).  

\subsection{ISPeL}
\label{s:ISPeL}
\begin{figure}[h]
\centerline{\includegraphics[width=\columnwidth]{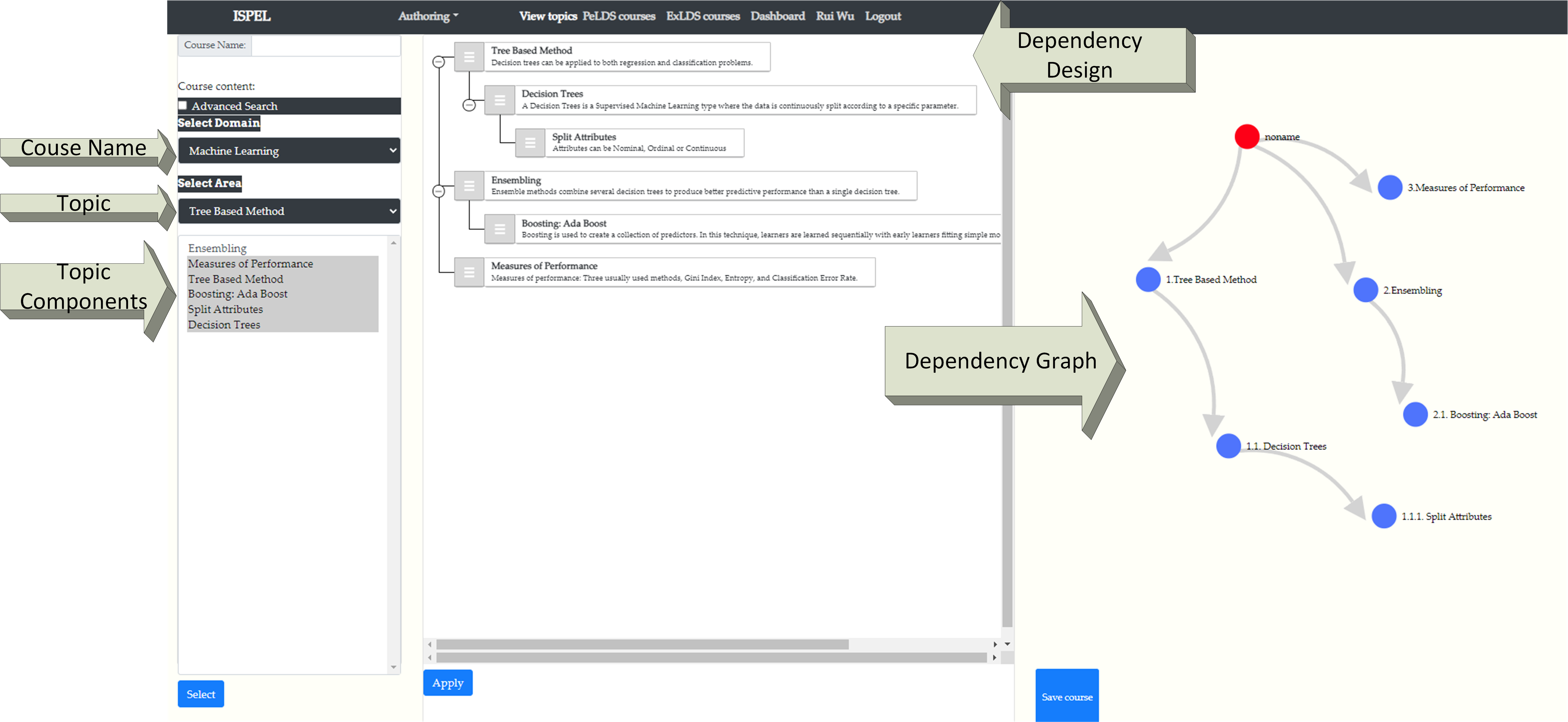}}
\caption{An instructor can choose topics (i.e., select area, see left top corner), select topic components, and design how to connect topic components (see middle column) with a book chapter and sub-chapter style to define the dependencies. When the “Apply” button is clicked, a dependency graph will be generated automatically.}
\label{fig:dependency_design}
\end{figure}

Interactive System for Personalized Learning (ISPeL)~\cite{ispel_website} is implemented based on the proposed personalized learning framework illustrated in this paper. The latest version of the ISPeL platform is hosted on an ECU server. When a student logs in, they can view course content (videos and images embedded in HTML files using responsive design able to be easily viewed on many types of devices) and check their learning progress. This design allows a student to learn course materials on different types of devices using a browser without installing other software. The instructor can upload course content to the ISPeL platform or customize and reuse topic components from another course. The instructor is able to organize course content using a dependency graph that is created automatically after an instructor orders the topic components in a book chapter/sub-chapter style (Fig.~\ref{fig:dependency_design}) with mouse button clicks and drags. The dependency graph will be generated based on the dependency hierarchy defined by the instructor.

\subsection{How to use our proposed method}
A community college or a university can obtain course content from a personalized learning server called ISPeL (see details in Section~\ref{s:ISPeL}) remotely and offer the robotics course on campus without hiring robotics experts. Students are required to complete a certain percentage of course content in each class, but the learning path can be different. In other words, each student can learn different topics in the same classroom by walking through the corresponding course materials, such as videos and sample code wrapped in Jupyter-notebooks,  a computer science language execution software. If a student has any questions, he/she should ask the general instructor first. The chatbot can answer basic questions, such as the robotics concept and syllabus-related questions. If the general instructor is not capable of answering the question, the instructor will check “frequently asked questions and guideline” document (including possible answers and steps to solve the problems) created by the course module designer and course content advisory committee. If the question still cannot be solved, the instructor will email the course module designer and course content advisory committee for consultancy. In our proposed framework, the main jobs of an instructor are to make sure students reach milestones in every class, collect students’ questions, and distribute back answers from the course module designer and course content advisory committee. We are inspired by the self-determination theory~\cite{deci2012self}, which shows increasing students’ autonomy can enhance their motivation and engagement. In principle, we believe it is important to ask students to show up in the classroom instead of completing the course content remotely. This is because in an “in-person” classroom an instructor can better ensure students make progress and manage devices correctly and students can discuss with each other to fill some knowledge gaps.

%------------------------------------------------

\section{User Study Results and Discussion}
\label{s:user_study}
To support the development of personalized learning framework we conducted two surveys. One was a user study survey with students taking a machine learning course (Fall 2022) where ISPeL was used as one method of course delivery. The user study survey was designed to assess how students used ISPeL as well as their perceptions of its quality and usefulness, including how ISPeL compared to other forms of instructional delivery, particularly PowerPoint. The survey also included open-ended items for students to make improvement recommendations. 

For the second survey, we collected student feedback from a two-week mini-course embedded within a one-semester course on special topics in engineering, in which robotics topics were delivered via ISPeL~\cite{shill2023wip}. The survey for the mini-courses was designed to gather student feedback on their course experiences in general~\cite{general_course} and its effect on their interest and motivation in robotics and future career plans. Because the study is inspired by Self-Determination Theory~\cite{deci2012self}, we adapted course evaluation items that have been used in previous studies with a focus on student autonomy~\cite{heinrich2021exploring,sheldon2012balanced}. Here we present findings on the course evaluation items only, given that only a selection of redesigned topics were piloted using ISPeL during a two-week period in an otherwise traditional one-semester course. We pooled student survey results across two semesters (Spring 2023 and Fall 2023) for reporting.

Results from the user study survey administered in the machine learning course indicate that students had favorable opinions of ISPeL in terms of organization and presentation of content and saw it as a useful tool to supplement their learning, particularly outside of class. In addition, students made several recommendations for improving ISPeL in terms of system design and content. 

% TODO read from here
Students responded to 11 items related to the perceived quality and usefulness of ISPeL (see Table~\ref{tbl:evaluation}). On a Likert scale, over 70\% of students selected agree or strongly agree about the quality and usefulness of ISPeL on 6 of 11 items. The most positively held views were on the overall usefulness of ISPeL to learn main concepts or topics, organization of content, and usefulness to prepare for other courses. Fewer students agreed that the layout of ISPeL was intuitive or that Jupyter notebook, a learning tool for coding used in the course, worked well with the system. Also, slightly over half of students agreed or strongly agreed with the statement that they wished other instructors used ISPeL, with over one-third neither agreeing nor disagreeing with the statement. Importantly, across items few students disagreed or strongly disagreed with statements about the quality or usefulness of ISPeL. 

% TODO add caption
\begin{table}[]
\centering
\caption{Students' Overall Evaluations of ISPeL}
\label{tbl:evaluation} 
\begin{adjustbox}{max width=\textwidth}
\begin{tabular}{|l|l|l|l|l|l|l|l|}
\hline
\textit{\textbf{Item}}                                                                                                                                                                    & \textit{\textbf{N}} & \textit{\textbf{\begin{tabular}[c]{@{}l@{}}Strongly   \\ Disagree\end{tabular}}} & \textit{\textbf{Disagree}} & \textit{\textbf{Neither}} & \textit{\textbf{Agree}} & \textit{\textbf{\begin{tabular}[c]{@{}l@{}}Strongly \\   Agree\end{tabular}}} & \textit{\textbf{\begin{tabular}[c]{@{}l@{}}Agree \\  + Strongly \\ Agree\end{tabular}}} \\ \hline
\begin{tabular}[c]{@{}l@{}}The ISPeL system is promising to help me learn \\ the main concepts in this class.\end{tabular}                                                                & 110                 & 1.8\%                                                                            & 2.7\%                      & 11.8\%                    & 53.6\%                  & 30.0\%                                                                        & 83.6\%                                                                                  \\ \hline
\begin{tabular}[c]{@{}l@{}}I believe the content of the ISPeL lesson is \\ helping me understand the topic (or would have helped\\  me understand the topic).\end{tabular}                & 109                 & 1.8\%                                                                            & 3.7\%                      & 14.7\%                    & 55.0\%                  & 24.8\%                                                                        & 79.8\%                                                                                  \\ \hline
\begin{tabular}[c]{@{}l@{}}Content for each topic on the ISPeL system \\ is clearly presented.\end{tabular}                                                                               & 109                 & 1.8\%                                                                            & 5.5\%                      & 13.8\%                    & 46.8\%                  & 32.1\%                                                                        & 78.9\%                                                                                  \\ \hline
\begin{tabular}[c]{@{}l@{}}The ISPeL system is promising to help me gain \\ skills that will be useful to me in other courses.\end{tabular}                                               & 110                 & 1.8\%                                                                            & 2.7\%                      & 18.2\%                    & 46.4\%                  & 30.9\%                                                                        & 77.3\%                                                                                  \\ \hline
\begin{tabular}[c]{@{}l@{}}The ISPeL system is a good resource when I need  \\  more information about a concept in this class.\end{tabular}                                              & 108                 & 4.6\%                                                                            & 2.8\%                      & 19.4\%                    & 41.7\%                  & 31.5\%                                                                        & 73.2\%                                                                                  \\ \hline
The topics on the ISPeL system are well   organized.                                                                                                                                      & 109                 & 1.8\%                                                                            & 3.7\%                      & 22.0\%                    & 38.5\%                  & 33.9\%                                                                        & 72.4\%                                                                                  \\ \hline
\begin{tabular}[c]{@{}l@{}}I believe the ISPeL system is helping me study\\  and prepare for my exams.\end{tabular}                                                                       & 110                 & 3.6\%                                                                            & 3.6\%                      & 31.8\%                    & 38.2\%                  & 22.7\%                                                                        & 60.9\%                                                                                  \\ \hline
\begin{tabular}[c]{@{}l@{}}The end of section quiz(s) {[}if present{]} is/are  \\  relevant to the topic(s) involved and is helping me \\ understand how to apply the topic.\end{tabular} & 107                 & 2.8\%                                                                            & 1.9\%                      & 36.4\%                    & 43.0\%                  & 15.9\%                                                                        & 58.9\%                                                                                  \\ \hline
\begin{tabular}[c]{@{}l@{}}The ISPeL system layout is intuitive to understand  \\  and easy to navigate.\end{tabular}                                                                     & 110                 & 3.6\%                                                                            & 10.0\%                     & 29.1\%                    & 35.5\%                  & 21.8\%                                                                        & 57.3\%                                                                                  \\ \hline
I wish more instructors used the ISPeL system.                                                                                                                                            & 108                 & 2.8\%                                                                            & 4.6\%                      & 37.0\%                    & 32.4\%                  & 23.1\%                                                                        & 55.5\%                                                                                  \\ \hline
Jupyter notebook works well with the ISPeL system.                                                                                                                                        & 110                 & 2.7\%                                                                            & 6.4\%                      & 40.9\%                    & 26.4\%                  & 23.6\%                                                                        & 50.0\%                                                                                  \\ \hline
\end{tabular}
\end{adjustbox}
\end{table}

Students were also asked to compare ISPeL to PowerPoint. In general, most students preferred ISPeL (42.1\%) to PowerPoint (27.1\%), with 30.8\% having no preference (see Table~\ref{tbl:evaluation}). In specific situations, students preferred ISPeL over PowerPoint on six of eight items presented to them, with the greatest preference for ISPeL in contexts where students want to make connections between topics, as a tool for additional information, and as an independent study aid. PowerPoint was only preferred as a tool for instructors to present course content and to organize the lesson. This pattern of results suggests that students prefer ISPeL as a tool for more personalized learning opportunities. 

\section{Conclusion and Future Work}
In this paper, we propose a student-centered personalized learning framework for undergraduate robotics education with general instructors. A learning platform called ISPeL is implemented based on this framework, and we have conducted a user study to collect students' feedback. The user study results show that our proposed method and ISPeL are promising in enhancing students' education, particularly in understanding the connections between topics and mastering the subjects.

% future work
For future data collection, we will use the second survey introduced in the User Study Results and Discussion Section. It will be used in future one-semester robotics courses (a traditional course and two redesigned courses) using a pre-post design. Students will answer motivation-related questions at the start of the course and both motivation and course evaluation questions at the end. The same survey will be administered pre and post in all courses to compare students’ classroom experiences between the traditional and redesigned courses in terms of course evaluations and changes in motivational beliefs and career plans.
% disable students
\ruic{In addition to data collection, our ongoing efforts involve enhancing our proposed framework to prioritize accessibility. We recognize the importance of accommodating disabled students, particularly those with visual or auditory impairments. Accordingly, adjustments will be implemented to ensure inclusivity and provide equal learning opportunities for all students, irrespective of their abilities or challenges. This commitment to accessibility underscores our dedication to fostering an inclusive educational environment that caters to the diverse needs of every learner.}

%\section*{Acknowledgements}
%This material is based in part upon work supported by: The National Science Foundation under grant number(s) NSF awards 
%    2142428, % ECU IUSE number
%    2142360, % UNR IUSE number
%    OIA-2019609, and % T2 Tic
%    OIA-2148788. % T1 Fire
%Any opinions, findings, and conclusions or recommendations expressed in this material are those of the author(s) and do not necessarily reflect the views of the National Science Foundation.

% During review, this is NOT allowed; Abstracts and papers are double-blind reviewed. It is the author’s responsibility to ensure that the requirements for double-blind review are met. The abstract and subsequent drafts should NOT include authors’ names or institutional affiliations nor should author names be in the file name or in document properties. It is not necessary to include references in the abstract. Be sure to indicate that your abstract is for the Community Engagement Division.

%----------------------------------------------------------------------------------------
%  REFERENCE LIST
%----------------------------------------------------------------------------------------
\vspace{4\baselineskip}\vspace{-\parskip} % Creaters proper 4 blank line spacing.
\footnotesize % Makes bibliography 10 pt font.
\bibliographystyle{unsrtnat} %Can use a different style as long as it is one which uses numbered references in the text.
\bibliography{ASEEpaper}

%----------------------------------------------------------------------------------------

\end{document}